\documentclass[10pt, a4paper]{article}
\usepackage{lrec}
\usepackage{graphicx}
\usepackage{tabularx}
\usepackage{color }
\usepackage{epstopdf}
\usepackage[latin1]{inputenc}

\usepackage{hyperref}

\newcommand{\SB}[1]{{\bf [\textcolor{green}{STEFANO #1}{\bf ]}}}

\newcommand{\secref}[1]{\StrSubstitute{\getrefnumber{#1}}{ }{ }}

\newcommand{\COMMENT}[1]{}

\title{\textbf{TLT-school: a Corpus of Non Native Children Speech}}

\name{R. Gretter, M. Matassoni, S. Bann\`o, D. Falavigna}

\address{Fondazione Bruno Kessler (FBK) \\
         Trento, Italy \\
         \{gretter, matasso, sbanno, falavi\}@fbk.eu\\}

\abstract{
This paper describes ``TLT-school'' a corpus of speech utterances collected in schools of northern Italy for assessing the performance of students learning both English and German. The corpus was recorded in the years 2017 and 2018 from students aged between nine and sixteen years, attending primary, middle and high school. 
All utterances have been scored, in terms of some predefined proficiency indicators, by human experts. In addition, most of utterances recorded in 2017 have been manually transcribed carefully. Guidelines and procedures used for manual transcriptions of utterances will be described in detail, as well as results achieved by means of an automatic speech recognition system developed by us. Part of the corpus is going to be freely distributed to scientific community particularly interested both in non-native speech recognition and automatic assessment of second language proficiency.
 \\ \newline
 }

\begin{document}

\maketitleabstract

\section{Introduction}

We have acquired large sets of both written and spoken data during the implementation of campaigns aimed at assessing the proficiency, at school, of Italian pupils learning both German and English. Part of the acquired data has been included in a corpus, named "Trentino Language Testing" in schools (TLT-school), that will be described in the following.

All the collected sentences have been annotated by human experts in terms
of some predefined ``indicators'' which, in turn, were used to assign the proficiency level to each student undertaking the assigned test. This level is expressed 
according to the well-known Common European Framework of Reference for Languages (Council of Europe, 2001) scale. The CEFR defines $6$ levels of proficiency: A1 (beginner), A2, B1, B2, C1 and C2. The levels considered in the evaluation campaigns where the data have been collected are: A1, A2 and B1. 


The indicators measure the linguistic competence of test takers both in relation to the content (e.g.\ grammatical correctness, lexical richness, semantic coherence, etc.)  and to the speaking capabilities (e.g.\ pronunciation, fluency, etc.).  Refer to Section~\ref{sec:acq} for a description of the adopted indicators. 

The learners are Italian students, between 9 and 16 years old. They took proficiency tests by answering question prompts provided in written form. The ``TLT-school'' corpus, that we are going to make publicly available, contains part of the spoken answers (together with the respective manual transcriptions) recorded during some of the above mentioned evaluation campaigns.  We will release the written answers in future. Details and critical issues found during the acquisition of the answers of the test takers will be discussed in Section~\ref{sec:acq}.

The tasks that can be addressed by using the corpus are very challenging and pose many problems, which have only partially been solved by the interested scientific community. 

From the ASR perspective, major difficulties are represented by: {\it a)} 
recognition of both child and non-native speech, i.e.\ Italian pupils speaking both English and German, {\it b)} presence of a large number of spontaneous speech phenomena (hesitations, false starts, fragments of words, etc.), {\it c)} presence of multiple languages (English, Italian and German words are frequently uttered in response to a single question), {\it d)} presence of a significant level of background noise due to the fact that the microphone remains open for a fixed time interval (e.g.\ 20 seconds - depending on the question), and {\it e)} presence of non-collaborative speakers (students often joke, laugh, speak softly, etc.). Refer to Section~\ref{sec:spoken}
for a detailed description of the collected spoken data
set. 

Furthermore, since the sets of data from which ``TLT-school'' was derived were primarily acquired for measuring proficiency of second language (L2)  learners, it is quite obvious to exploit the corpus for automatic speech rating. To this purpose, one can try to develop automatic approaches to reliably estimate the above-mentioned indicators used by the human experts who scored the answers of the pupils (such an approach is described in \cite{icassp2019}). However, it has to be noticed that scientific literature  proposes to use several features and indicators for automatic speech scoring, partly different  from those adopted in ``TLT-school'' corpus (see below for a brief review of the literature). Hence, we believe that adding new annotations to the corpus, related to particular aspects of language proficiency, can stimulate research and experimentation in this area. 

\COMMENT{ 
that can be addressed using the ``TLT-school'' corpus, which was also the main purpose for developing the corpus itself,  is automatic rating of the proficiency levels of students who undertook the tests. 
First of all it is not easy to define which are the most suitable ``indicators'' to be used for estimating the CEFR levels and, then, it has to be defined how to weigh them to compose the final  judgements. Note that the indicators  adopted in the evaluation campaigns mentioned above are equally weighted to give the CEFR level and do not allow to consider important aspects of the proficiency of L2 learners \SB{Stefano si puo` accennare  qualcosa in piu` circa i limiti relativi agli indicatori che usa IPRASE? Da specificare meglio nella section conclusions and future works}. Finally, it has to be addressed the problem of achieving reliable estimates of the indicators given the collected data.
}

Finally, it is worth mentioning that also written responses of ``TLT-school'' corpus are characterised by a high level of noise due to: spelling errors, insertion of word fragments, presence of words belonging to multiple languages, presence of off-topic answers (e.g. containing jokes, comments not related to the questions, etc.). This set of text data will allow scientists to investigate both language and behaviour of pupils learning second languages at school.  
Written data are described in detail in Section~\ref{sec:written}


{\bf Relation to prior work.} Scientific literature is rich in approaches for automated assessment of spoken language proficiency. Performance is directly dependent on ASR accuracy which, in turn, depends on the type of input, read or spontaneous, and on the speakers' age, adults or children (see  \cite{eskenazi2009} for an overview of spoken language technology for education). A recent publication reporting an overview of state-of-the-art automated speech scoring technology as it is currently used at Educational Testing Service (ETS) can be found in \cite{zechner2019}.


In order to address automatic assessment of complex spoken tasks requiring more general communication capabilities from L2 learners, the AZELLA data set  \cite{cheng2014}, developed by Pearson, has been collected and used as benchmark for some researches~\cite{angeliki2014,cheng2014}. The corpus contains $1,500$ spoken tests, each double graded by human professionals, from a variety of tasks. 



A public set of spoken data has been recently distributed in a spoken CALL (Computer Assisted Language Learning) shared task\footnote{https://regulus.unige.ch/spokencallsharedtask\_3rdedition/ for details.}  
where Swiss students learning English had to answer to both written and spoken prompts. The goal of this challenge is to label students' spoken responses  as ``accept''  or ``reject''. Refer to \cite{Baur2018} for details of the challenge and of the associated data sets.

Many non-native speech corpora (mostly in English as target language) have been collected during the years. A list, though not recent, as well as a brief description of most of them can be found in \cite{noeth2007}. The same paper also gives information on how the data sets are distributed and can be accessed (many of them are available through both LDC\footnote{https://www.ldc.upenn.edu/}  and ELDA\footnote{http://www.elra.info/en/about/elda/} agencies). Some of the corpora also provide proficiency ratings to be used in CALL
applications. Among them, we mention the ISLE corpus \cite{menzel2000isle}, which also contains transcriptions at the phonetic level and was used in the experiments reported in~\cite{icassp2019}. 

Note that all corpora mentioned in \cite{noeth2007} come from adult speech while, to our knowledge, the access to publicly available non-native children's speech corpora, as well as of children's speech corpora in general, is still scarce. 
Specifically concerning non-native children's speech, we believe worth mentioning the following corpora. The PF-STAR corpus (see \cite{batliner2005}) contains English utterances read by both Italian and German children, between 6 and 13 years old. The same corpus also contains utterances read by English children. The {\em ChildIt} corpus \cite{russell2007} contains English utterances (both read and imitated) by Italian children. 

By distributing ``TLT-school'' corpus, we hope to help researchers to investigate novel approaches and models in the areas of both non-native and children's speech and to build related benchmarks.

\section{Data Acquisition}
\label{sec:acq}

In Trentino, an autonomous region in northern Italy, there is a series of evaluation campaigns underway for testing L2 linguistic competence of Italian students taking proficiency tests in both English and German. A set of three evaluation campaigns is underway, two having been completed in 2016 and 2018, and a final one scheduled in 2020. 
Note that the ``TLT-school'' corpus refers to only the 2018 campaign, that was split in two parts: 2017 try-out data set (involving about 500 pupils) and the actual 2018 data (about 2500 pupils).
Each of the three campaigns (i.e. 2016, 2018 and 2020) involves about 3000 students ranging from 9 to 16 years, belonging to four different school grade levels and three proficiency levels (A1, A2, B1). The schools involved in the evaluations are located in most part of the Trentino region, not only in its main towns; 
Table~\ref{tab:plan} highlights some information about the pupils that took part to the campaigns. Several tests, aimed at assessing the language learning skills of the students, were carried out by means of multiple-choice questions, which can be evaluated automatically. However, a detailed linguistic evaluation cannot be performed without allowing the students to express themselves in both written sentences and spoken utterances, which typically require the intervention of human experts to be scored.

\begin{table}[tb]
\caption{Evaluation of L2 linguistic competences in Trentino: level, grade, age and number of pupils participating in the evaluation campaigns. Most of the pupils did both the English and the German tests.}
\label{tab:plan}
\centering
\begin{tabular}{ l c c r r r }
\\ \hline
CEFR  & Grade, School & Age & \multicolumn{3}{c}{Number of pupils} \\ 
     &                 &       & 2016 & 2017 & 2018 \\ \hline
 A1  & 5, primary      &  9-10 & 1074 &  320 &  517 \\ 
 A2  & 8, secondary    & 12-13 & 1521 &  111 &  614 \\ 
 B1  & 10, high school & 14-15 &  378 &  124 & 1112 \\ 
 B1  & 11, high school & 15-16 &  141 &    0 &  467 \\ \hline
 tot &   5-11          &  9-16 & 3114 &  555 & 2710 \\ \hline 
\end{tabular}
\end{table}



\begin{table}[tb]
\caption{Written data collected during different evaluation campaigns. Column ``\#Q'' indicates the total number of different (written) questions presented to the pupils.}
\label{tab:datawritten}
\centering
\begin{tabular}{ l c r r r r }
\\
\hline
 Year & Lang & \#Pupils  & \#Sentences   & \#Tokens  & \#Q  \\ \hline
 2016 & ENG & 3074 & 5062 & 299138 & 20 \\ 
 2016 & GER & 2870 & 4658 & 192144 & 25 \\ 
 2017 & ENG &  533 &  758 &  37225 &  5 \\ 
 2017 & GER &  529 &  745 &  30802 &  5 \\ 
 2018 & ENG & 2560 & 4600 & 293958 &  5 \\ 
 2018 & GER & 2200 & 3889 & 202309 &  5 \\ \hline
\end{tabular}
\end{table}

\begin{table}[tb]
\caption{Spoken data collected during different evaluation campaigns. Column ``\#Q'' indicates the total number of different (written) questions presented to the pupils.}
\label{tab:dataspeech}
\centering
\begin{tabular}{ l c r r r r }
\\
\hline
 Year &Lang & \#Pupils   & \#Utterances    & Duration  & \#Q  \\ \hline
 2016 & ENG & 2748 & 17462 & 69:03:37 &  85 \\
 2016 & GER & 2542 & 15866 & 60:03:01 & 101 \\ 
 2017 & ENG &  511 &  4112 & 16:25:45 &  24 \\ 
 2017 & GER &  478 &  3739 & 15:33:06 &  23 \\ 
 2018 & ENG & 2332 & 15770 & 93:14:53 &  24 \\ 
 2018 & GER & 2072 & 13658 & 95:54:56 &  23 \\ \hline
 \hline 
\end{tabular}
\end{table}

\begin{table}[tb]
\caption{List of the indicators used by human experts to evaluate specific linguistic competences.}
\label{tab:indicators}
\centering
\begin{tabular}{ p{7.5cm} }
\\
\hline
{\bf lexical richness}: lexical properties, lexical appropriateness \\
\hline
{\bf pronunciation and fluency}: pronunciation, fluency, discourse pronunciation and fluency \\
\hline
{\bf syntactical correctness}: correctness and formal competences, morpho-syntactical correctness, orthography  and punctuation \\
\hline
{\bf fulfillment on delivery}: fulfillment of the task, relevancy of the answer \\
\hline
{\bf coherence and cohesion}: coherence and cohesion, general impression \\
\hline
{\bf communicative, descriptive, narrative skills}: communicative efficacy, argumentative abilities, descriptive abilities, abilities to describe one's own feelings, etc. \\
\hline
\end{tabular}
\end{table}

\begin{table*}[tb]
\caption{Samples of written answers to English questions. On each line the CEFR proficiency level, the question and the answer are reported. Other information (session and question id, total and individual scores, school/class/student anonymous id) is also available but not included below.}
\vspace{0.2cm}
\label{tab:writtensamples}
\small
\centering
\begin{tabular}{cp{6cm}p{9cm}}
\hline
CEFR & Question & Answer \\
\hline
A1 &
You are on a trip to Trentino with your family. Add a message to a picture you took and want to send it to a friend. Tell us: 1. where you are; 2.what you do; 3. what you like or dislike. & 
dear tiago . i'm swimming in the lake . there are some beautiful mountains . i swim in the levico lake . later i go home on the bikeand i eats ice creamwith my brother and my father . i like water is beautiful but i don't like the sun is very very hot ! is very impressive levico lake . see you soon . byee ! kacper 

------------------------------------------------------

hello , i'm in the lake with my family . i play football with my dad and i eat a ice cream . the water is beautiful . it's very sunny . goodbye see you soon . \\
\hline
A2 & 
Reply to Susan

Hi! How are you? I've just received a new tennis racket. Would you like to meet at the sports centre and play a little? We can play for an hour and then we can get an ice cream together. Can you come at 5 o'clock? Don't forget to bring your tennis shoes, ok? I'm really looking forward to playing with you! Bye, Susan. &
hello susan ! i'm fine thanks and you ? i'm very happy for you and for your message and i would like see your new racket but unfortunately today i can't come . tomorrow i have a very important football match and i must wake up at six o'clock so i need to sleep more than usually . we can meet  

------------------------------------------------------

hello susan i'm fine . i'm sorry but i can't come with you , because i go to land between london for one concert at five o'clock . bye .  \\
\hline
B1 &
Write an English post for your blog where you talk about what you need to do to learn a language well. &
if you want to learn a new language you can make it . the first thing is studying many words and make a course . the second thing is go out of your state and arrive at the state where speak the language that you want learn . the first days are more difficult , but then its more easy . you must study the grammar too .  \\
\hline
B1 &
Write a short email to a friend of yours to tell him / her that you intend to start studying another foreign language and what the reasons are. &
hi sophie ! how are you ? i am writing to you because i desire to say you that i will start to study a new langueges . why i decide it ? because i wont to live in spain in the future . what do you thing ? i wait your answer . with love ! bye !   \\ 
\hline
\end{tabular}
\end{table*}

Tables~\ref{tab:datawritten} and \ref{tab:dataspeech} report some statistics extracted from both the written and spoken data collected so far in all the campaigns. Each written or spoken item received a total score by human experts, computed by summing up the scores related to $6$ indicators in 2017/2018 (from $3$ to $6$ in the 2016 campaign, according to the proficiency levels and the type of test).
Each indicator can assume a value 0, 1, 2, corresponding to 
bad, medium, good, respectively.


The list of the indicators used by the experts to score written sentences and spoken utterances in the evaluations, grouped by similarity, is reported in Table~\ref{tab:indicators}. 
Since every utterance was scored by only one expert, it was not possible to evaluate any kind of agreement among experts. For future evaluations, more experts are expected to provide independent scoring on same data sets, so this kind of evaluation will be possible.


\subsection{Prompts}
The speaking part of the proficiency tests in 2017/2018 consists of 47 question prompts provided in written form: 24 in English and 23 in German, divided according to CEFR levels. Apart from A1 level, which differs in the number of questions (11 for English; 10 for German), both English and German A2 and B1 levels have respectively 6 and 7 questions each.
As for A1 level, the first four introductory questions are the same ({\it How old are you?}, {\it Where do you live?}, {\it What are your hobbies?}, {\it Wie alt bist du?}, {\it Wo wohnst du?}, {\it Was sind deine Hobbys?}) or slightly different ({\it What's your favourite pet?}, {\it Welche Tiere magst du?}) in both languages, whereas the second part of the test puts the test-takers in the role of a customer in a pizzeria (English) or in a bar (German).

\COMMENT{
Highlighting the difference between the two environments might seem nitpicking, but this aspect together with the degree of openness of the questions has a strong and direct influence on the test-takers’ answers. For example the expected answers to ‘Would you like to order a pizza?’ might be either ‘Yes, please.’ or ‘No, thanks.’, while its German counterpart ‘Was m\"ochtest du essen?’ provides a much wider range of answers. It is no coincidence that ‘pizza’ is the fourth most frequent word in the English A1 test, but such word is not really worthy of interest from a phonetic and lexical point of view for obvious reasons. 
}
A2 level test is composed of small talk questions which relate to everyday life situations. In this case, questions are more open-ended than the aforementioned ones and allow the test-takers to interact by means of a broader range of answers.
Finally, as for B1 level, questions are similar to A2 ones, but they include a role-play activity in the final part, which allows a good amount of freedom and creativity in answering the question.
\COMMENT{
There is another issue that emerges from the answers of both A2 and B1 level tests: the question prompts do not require test takers to shift tense. This lack is quite remarkable not just for the impact that has on the data collection itself, but also on the performance of the ASR system.

The written part of the proficiency tests is made up of 5 questions for each language: 1 for A1, 2 for A2 and 2 for B1. The question prompt of the A1 level test is in Italian and asks the test-takers to write a letter to a friend, that should include a short descriptive text. 
A similar task is also required – again in Italian – in one of the two questions of A2 level, whereas the second question is prompted in the target language and is a short text message from a friend who invites the test-taker to go out and play sport. 
Likewise, the B1 test consists of two questions. Once again, one asks the test-takers to write a letter, while the other requires them to write a brief argumentative text. Similarly to the A2 test, the question prompts of the German B1 test are in Italian and German, whereas the respective ones of the English B1 test are only in Italian. The presence of both source and target languages in the question prompts is quite confusing and problematic.
}


\subsection{Written Data}
\label{sec:written}

Table~\ref{tab:datawritten} reports some statistics extracted from the written data collected so far. 
In this table, the number of pupils taking part in the English and German evaluation is reported, along with the number of sentences and tokens, identified as character sequences bounded by spaces.

It is worth mentioning that the collected texts contain a large quantity of errors of several types: orthographic, syntactic, code-switched words (i.e. words not in the required language), jokes, etc.
Hence, the original written sentences have been processed in order to produce ``cleaner'' versions, in order to make the data usable for some research purposes (e.g.\ to train language models, to extract features for proficiency assessment, \ldots). 


To do this, we have applied some text processing, that in sequence:

$\bullet$ removes strange characters;

$\bullet$ performs some text normalisation (lowercase, umlaut, numbers, \ldots) and tokenisation (punctuation, etc.)

$\bullet$ removes / corrects non words (e.g.\ {\it hallooooooooooo} becomes {\it hallo}; {\it aaaaaaaaeeeeeeeeiiiiiiii} is removed)

$\bullet$ identifies the language of each word, choosing among Italian, English, German;

$\bullet$ corrects common typing errors (e.g. {\it ai em} becomes {\it i am})

$\bullet$ replaces unknown words, with respect to a large lexicon, with the label {\it $<$unk$>$}.

Table~\ref{tab:writtensamples} reports some samples of written answers.

\subsection{Spoken Data}
\label{sec:spoken}
Table~\ref{tab:dataspeech} reports some statistics extracted from the acquired spoken data. 
Speech was recorded in classrooms, whose equipment depended on each school. In general, around 20 students took the test together, at the same time and in the same classrooms, so it is quite common that speech of mates or teachers often overlaps with the speech of the student speaking in her/his microphone. Also, the type of microphone depends on the equipment of the school. On average, the audio signal quality is nearly good, while the main problem is caused by a high percentage of extraneous speech.
This is due to the fact that organisers decided to use a fixed duration - which depends on the question - for recording spoken utterances, so that all the recordings for a given question have the same length. However, while it is rare that a speaker has not enough time to answer, it is quite common that, especially after the end of the utterance, some other speech (e.g. comments, jokes with mates, indications from the teachers, etc.) is captured.
In addition, background noise is often present due to several sources (doors, steps, keyboard typing, background voices, street noises if the windows are open, etc). Finally,  it has to be pointed out that many answers are whispered and difficult to understand.


\section{Manual Transcriptions}
\label{sec:annotation}

In order to create both an adaptation and an evaluation set for ASR,
we manually transcribed part of the 2017 data sets. We defined an initial set of guidelines for the annotation, which were used by 5 researchers to manually transcribe about 20 minutes of audio data. This experience led to a discussion, from which a second set of guidelines originated, aiming at reaching a reasonable trade-off between transcription accuracy and speed. As a consequence, we decided to apply the following transcription rules:
\begin{itemize}
\item
only the main speaker has to be transcribed; presence of other voices (schoolmates, teacher) should be reported only with the label ``@voices'',
\item 
presence of whispered speech was found to be significant, so it should be explicitly marked with the label ``()'',
\item
badly pronounced words have to be marked by a ``\#'' sign, without trying to phonetically transcribe the pronounced sounds; ``\#*'' marks incomprehensible speech;
\item speech in a different language from the target language has to be reported by means of an explicit marker {\it ``I am 10 years old @it(io ho gi\`a risposto)''}.
\end{itemize}

\begin{table}[bt]
\caption{Inter-annotator agreement between pairs of students, in terms of words. Students transcribed English utterances first and German ones later.}
\label{tab:agreement}
\small
\centering
\begin{tabular}{ c c c c c }
\\
\hline
 High   & Language & \#Transcribed & \#Different & Agreement \\
 school &          & words         & words       &           \\
\hline
C & English  &  965 & 237 & 75.44\% \\
C & German   &  822 & 139 & 83.09\% \\
\hline
S & English & 1370 & 302 & 77.96\% \\
S & German  & 1290 & 226 & 82.48\% \\
\hline
\end{tabular}
\end{table}


\begin{table}[tb]
\caption{Statistics from the spoken data sets (2017) used for ASR.}
\label{tab:TestAsrV6}
\small
\centering
\begin{tabular}{ l | c | c c | c c } 
\\
\hline
id               & \# of  & \multicolumn{2}{c|}{duration} & \multicolumn{2}{c}{tokens} \\ 
                 &  utt.  & total    &    avg   & total   &    avg \\ \hline
 Ger Train All   &  1448  &  04:47:45 &  11.92  &  9878   &   6.82 \\  
 Ger Train Clean &   589  &  01:37:59 &   9.98  &  2317   &   3.93 \\ \hline
 Eng Train All   &  2301  &  09:03:30 &  14.17  & 26090   &  11.34 \\
 Eng Train Clean &   916  &  02:45:42 &  10.85  &  6249   &   6.82 \\ \hline \hline
 Ger Test All    &   671  &  02:19:10 &  12.44  &  5334   &   7.95 \\  
 Ger Test Clean  &   260  &  00:43:25 &  10.02  &  1163   &   4.47 \\ \hline
 Eng Test All    &  1142  &  04:29:43 &  14.17  & 13244   &  11.60 \\  
 Eng Test Clean  &   423  &  01:17:02 &  10.93  &  3404   &   8.05 \\ \hline
\end{tabular}
\end{table}

Next, we concatenated utterances to be transcribed into blocks of about 5 minutes each. We noticed that knowing the question and hearing several answers could be of great help for transcribing some poorly pronounced words or phrases. Therefore, each block contains only answers to the same question, explicitly reported at the beginning of the block. 

We engaged about 30 students from two Italian linguistic high schools (namely ``C'' and ``S'') to perform manual transcriptions.

After a joint training session,
we paired students together. Each pair first transcribed, individually, the same block of $5$ minutes. Then, they went through a comparison phase, where each pair of students discussed their choices and agreed on a single transcription for the assigned data.  Transcriptions made before the comparison phase were retained to evaluate inter-annotator agreement. 
Apart from this first 5 minute block, each utterance was transcribed by only one transcriber.
Inter-annotator agreement for the 5-minute blocks is shown in Table~\ref{tab:agreement} in terms of words (after removing hesitations and other labels related to background voices and noises, etc.). The low level of agreement reflects the difficulty of the task.

In order to assure quality of the manual transcriptions, every sentence transcribed by the high school students was automatically processed to find out possible formal errors, and manually validated by researchers in our lab.


Speakers were assigned either to training or evaluation sets, with proportions of $\frac{2}{3}$ and $\frac{1}{3}$, respectively; then training and evaluation lists were built, accordingly.
Table~\ref{tab:TestAsrV6} reports statistics from the spoken data set. The id {\em All} identifies the whole data set, while {\em Clean} defines the subset in which sentences containing background voices, incomprehensible speech and word fragments were excluded.

\section{Usage of the Data}
\label{sec:expe}

From the above description it appears that the corpus can be effectively used in many research directions.

\subsection{ASR-related Challenges}
\label{sec:asr}

The spoken corpus features non-native speech recordings in real classrooms and, consequently, peculiar phenomena appear and can be investigated. Phonological and cross-language interference requires specific approaches for accurate acoustic modelling. Moreover, for coping with cross-language interference it is important to consider alternative ways to represent specific words (e.g.\ words of two languages with the same graphemic representation).

Table~\ref{table:icassp_wer_results}, extracted from \cite{icassp2019}, reports WERs obtained on evaluation data sets with a strongly adapted ASR,  demonstrating the difficulty of the related speech recognition task for both languages. Refer to \cite{icassp2018} for comparisons with a different non-native children speech data set and to scientific literature \cite{WilJac96,DasNixPic98,LiRus01,GiuGer03,PotNar03,GerGiuBru07,GerGiuBru09,liao2015,serizel2016} for detailed descriptions of children speech recognition and related issues. Important, although not exhaustive of the topic, references on non-native speech recognition can be found in \cite{Wang2003a,Wang2003b,Oh2006,strik2009,Steidl2004,bouselmi2006,duan2017,li2016,lee2015,das2015}.

As for language models, accurate transcriptions of spoken responses demand for models able to cope with not well-formed expressions (due to students' grammatical errors).  Also the presence of code-switched words, words fragments and spontaneous speech phenomena requires specific investigations to reduce their impact on the final performance. 

We believe that the particular domain and set of data pave the way to investigate into various ASR topics, such as:  non-native speech, children speech, spontaneous speech, code-switching, multiple pronunciation, etc.  

\subsection{Data Annotation}
The corpus has been (partly) annotated using the guidelines presented in Section~\ref{sec:annotation} on the basis of a preliminary analysis of the most common acoustic phenomena appearing in the data sets. 

Additional annotations could be included to address topics related to other spurious segments, as for example: understandable words pronounced in other languages or by other students, detection of phonological interference, detection of spontaneous speech phenomena, detection of overlapped speech, etc.
In order to measure specific proficiency indicators, e.g. related to pronunciation and fluency, suprasegmental annotations can be also inserted in the corpus.

\subsection{Proficiency Assessment of L2 Learners}
The corpus is a valuable resource for training and evaluating a scoring classifier based on different approaches. 
Preliminary results \cite{icassp2019} show that the usage of suitable linguistic features mainly based on statistical language models allow to predict the scores assigned by the human experts. 

The evaluation campaign has been conceived to verify the expected proficiency level according to class grade; as a result, although the proposed test cannot be used to assign a precise score to a given student, it allows to study typical error patterns according to age and level of the students. 

Furthermore, the fine-grained annotation, at sentence level, of the  indicators described above is particularly suitable for creating a test bed for approaches based on ``word embeddings'' \cite{chen2018,oh2017,russell2019} to automatically estimate the language learner proficiency. Actually, the experiments reported in \cite{chen2018} demonstrate superior performance of word-embeddings for speech scoring with respect to the well known (feature-based) SpeechRater system \cite{zechner2009,zechner2019}.  In this regard, we believe that additional, specific annotations can be developed and included in the ``TLT-school'' corpus. 
 


\begin{table}[tb]
\caption{WER results on 2017 spoken test sets.}
\label{table:icassp_wer_results}
\centering
\begin{tabular}{ c c }
\\   \hline
 German & English \\ \hline 
42.6 &  35.9   \\ 
\hline
\end{tabular}
\end{table}

\begin{table}[tb]
\caption{Words suitable for pronunciation analysis. Data come from the 2017 manually transcribed data. Numbers indicate the number of occurrences, divided into test and training, with good and bad pronounciations.}
\label{tab:mispronwords}
\small
\centering
\begin{tabular}{ l | r  | c c | c c } 
\hline
      word  &  tot occ & \multicolumn{2}{c|}{good vs. bad} & \multicolumn{2}{c}{good vs. bad} \\ \hline  \hline
                   German &&         \multicolumn{2}{c|}{Test GER}       &\multicolumn{2}{c}{Train GER}     \\ \hline
             ich & 1132 &   317 &  32  &   735 &  48   \\
  lieblingsessen &  113 &    22 &  15  &    45 &  31   \\
             ist &  374 &   131 &   9  &   211 &  23   \\
            mein &  204 &    52 &   9  &   129 &  14   \\
       tsch\"uss &   33 &     6 &   5  &     7 &  15   \\
          h\"ore &   19 &     2 &   4  &     4 &   9   \\
             alt &  191 &    67 &   7  &   111 &   6   \\
   fr\"uhst\"uck &   31 &     4 &   3  &    15 &   9   \\
           milch &   22 &     1 &   6  &     9 &   6   \\
          heisse &   54 &    14 &   3  &    30 &   7   \\ \hline  \hline
          English    &&        \multicolumn{2}{c|}{Test ENG}     &  \multicolumn{2}{c}{Train ENG}      \\ \hline
       favourite &  578 &   171 &  17  &   362 &  28   \\
             pet &  169 &    49 &  10  &    96 &  14   \\
           thank &  179 &    57 &   4  &   102 &  16   \\
            live &  291 &    87 &   8  &   185 &  11   \\
      volleyball &   97 &    22 &   5  &    60 &  10   \\
        football &  246 &    75 &   7  &   157 &   7   \\
           years &  170 &    47 &   2  &   109 &  12   \\
         subject &   60 &    13 &   7  &    34 &   6   \\
          prefer &  116 &    37 &   6  &    66 &   7   \\
         friends &  120 &    27 &   3  &    82 &   8   \\ \hline
\end{tabular}

\end{table}

\subsection{Modelling Pronunciation}

By looking at the manual transcriptions, it is straightforward to detect the most problematic words, i.e.\ frequently occurring words, which were often marked as mispronounced (preceded by label ``\#''). This allows to prepare a set of data composed by good pronounced vs. bad pronounced words. 

A list of words, partly mispronounced, is shown in Table~\ref{tab:mispronwords}, from which one can try to model  typical pronunciation errors (note that other occurrences of the selected words could be easily extracted from the non-annotated data).  
\COMMENT{
Manually transcribed sentences were analysed to extract occurrences of
the same word (e.g.\ favourite or lieblingsessen) that, when  marked with a  "#", were
badly pronounced.  Data are divided into train and test.  
From this analysis, some table was obtained, see Table, that could be used:
Note that other occurrences of the selected words could be easily extracted from the non-annotated data.
Some further manual checking and annotation could be carried out to model typical pronunciation errors.
}
Finally, as mentioned above, further manual checking and annotation could be introduced to improve modelling of pronunciation errors.

\section{Distribution of the Corpus}

The corpus to be released is still under preparation, given the huge amount of spoken and written data; in particular, some checks are in progress in order to:
\begin{itemize}
\item remove from the data responses with personal or inadequate content  (e.g.\  bad language);
\item normalise the written responses (e.g.\  upper/lower case, punctuation, evident typos);
\item normalise and verify the consistency of the transcription of spoken responses;
\item check the available human scores and - if possible -  merge or map the scores according to more general performance categories (e.g.\  delivery, language use, topic development) and an acknowledged scale (e.g.\  from 0 to 4)\footnote{https://www.ets.org/s/toefl/pdf/toefl\_speaking\_rubrics.pdf}.
\end{itemize}

In particular, the proposal for an international challenge focused on non-native children speech recognition is being submitted where an English subset will be released and the perspective participants are invited to propose and evaluate state-of-art techniques for dealing with the multiple issues related to this challenging ASR scenario (acoustic and language models, non-native lexicon, noisy recordings, etc.).

\section{Conclusions and Future Works}

We have described ``TLT-school'', a corpus of both spoken and written answers collected during language evaluation campaigns carried out in schools of northern Italy. The procedure used for data acquisition and for their annotation in terms of proficiency indicators has been also reported. Part of the data has been manually transcribed according to some guidelines: this set of data is going to be made publicly available. 
\COMMENT{A new campaign is going to be conducted in 2020, with which we hope to increase the size of the corpus. At the same time, we aim at augmenting the annotations included in the corpus in order to cope with tasks more related to the L2 learning research area. To this purpose we also hope 
to arouse sufficient interest from the interested scientific community.}
With regard to data acquisition, some limitations of the corpus have been observed that might be easily overcome during next campaigns. Special attention should be paid to enhancing the elicitation techniques, starting from adjusting the questions presented to test-takers. Some of the question prompts show some lacks that can be filled in without major difficulty: on the one hand, in the spoken part, questions do not require test-takers to shift tense and some are too suggestive and close-ended; on the other hand, in the written part, some question prompts are presented both in source and target language, thus causing or encouraging code-mixing and negative transfer phenomena. The elicitation techniques in a broader sense will be object of revision (see \cite{cooke1994} and specifically on children speech \cite{beckman2017}) in order to maximise the quality of the corpus.
As for proficiency indicators, one first step that could be taken in order to increase accuracy in the evaluation phase both for human and automatic scoring would be to divide the second indicator (pronunciation and fluency) into two different indicators, since fluent students might not necessarily have good pronunciation skills and vice versa, drawing for example on the IELTS~\footnote{https://www.ielts.org} Speaking band descriptors.
Also, next campaigns  might consider an additional indicator specifically addressed to score prosody (in particular intonation and rhythm), especially for A2 and B1 level test-takers.
Considering the scope of the evaluation campaign, it is important to be aware of the limitations of the associated data sets: proficiency levels limited to A1, B1 and B2 (CEFR); custom indicators conceived for expert evaluation (not particularly suitable for automated evaluation); limited amount of responses per speaker. Nevertheless, as already discussed, the fact that the TLT campaign was carried out in 2016 and 2018 in the whole Trentino region
makes the corpus a valuable linguistic resource for a number of studies associated to second language acquisition and evaluation. In particular, besides the already introduced proposal for an ASR challenge in 2020, other initiatives for the international community can be envisaged: a study of a fully-automated evaluation procedure without the need of experts' supervision; the investigation of end-to-end classifiers that directly use the spoken response as input and produce proficiency scores according to suitable rubrics. 

\section{Acknowledgements}
This work has been partially funded by IPRASE (http://www.iprase.tn.it) under the project ``TLT -  Trentino Language Testing 2018''. We thank ISIT (http://www.isit.tn.it) for having provided the data and the reference scores.

\COMMENT{

\subsection{Bibliographical References}
Bibliographical references should be listed in alphabetical order at the
end of the article. The title of the section, ``Bibliographical References'',
should be a level 1 heading. The first line of each bibliographical reference
should be justified to the left of the column, and the rest of the entry should
be indented by 0.35 cm.

The examples provided in Section \secref{main:ref} (some of which are fictitious
references) illustrate the basic format required for articles in conference
proceedings, books, journal articles, PhD theses, and chapters of books.

\subsection{Language Resource References}

Language resource references should be listed in alphabetical order at the end
of the article.

\section*{Appendix: How to Produce the \texttt{.pdf} Version}

In order to generate a PDF file out of the LaTeX file herein, when citing
language resources, the following steps need to be performed:

\begin{itemize}
    \item{Compile the \texttt{.tex} file once}
    \item{Invoke \texttt{bibtex} on the eponymous \texttt{.aux} file}
    \item{Compile the \texttt{.tex} file twice}
\end{itemize}
}

\section{Bibliographical References}
\label{main:ref}

\bibliographystyle{lrec}

\bibliography{refs}

\begin{thebibliography}{}

\bibitem[\protect\citename{Angeliki and Cheng}2014]{angeliki2014}
Angeliki, M. and Cheng, J.
\newblock (2014).
\newblock Using {Deep} {Neural} {Networks} to improve proficiency assessment
  for children {English} language learners.
\newblock In {\em Proc. of Interspeech}, pages 1468--1472.

\bibitem[\protect\citename{Batliner \bgroup et al.\egroup }2005]{batliner2005}
Batliner, A., Blomberg, M., D'Arcy, S., Elenius, D., Giuliani, D., Gerosa, M.,
  Hacker, C., Russell, M., Steidl, S., and Wong, M.
\newblock (2005).
\newblock The {PF-STAR} children's speech corpus.
\newblock In {\em Proc. of Eurospeech}, pages 2761--2764.

\bibitem[\protect\citename{Baur \bgroup et al.\egroup }2018]{Baur2018}
Baur, C., Caines, A., Chua, C., Gerlach, J., Qian, M., Rayner, M., Russell, M.,
  Strik, H., and Wei, X.
\newblock (2018).
\newblock Overview of the 2018 spoken call shared task.
\newblock In {\em Proc. of Interspeech}, pages 2354--2358, Hyderabad, India.

\bibitem[\protect\citename{Beckman \bgroup et al.\egroup }2017]{beckman2017}
Beckman, M., Plummer, A., Munson, B., and Reidy, P.~F.
\newblock (2017).
\newblock Methods for eliciting, annotating, and analyzing databases for
  childspeech development.
\newblock {\em Computer Speech and Language}, (45):278--299.

\bibitem[\protect\citename{Bouselmi \bgroup et al.\egroup }2006]{bouselmi2006}
Bouselmi, G., Fohr, D., Illina, I., and Haton, J.~P.
\newblock (2006).
\newblock Multilingual non-native speech recognition using phonetic
  confusion-based acoustic model modification and graphemic constraints.
\newblock In {\em Proc. of ICSLP}, pages 109--112.

\bibitem[\protect\citename{Chen \bgroup et al.\egroup }2018]{chen2018}
Chen, L., Tao, J., Ghaffarzadegan, S., and Qian, Y.
\newblock (2018).
\newblock End-to-end neural network based automated speech scoring.
\newblock In {\em {Proc. of ICASSP}}, pages 6234--6238, Calgary, Canada.

\bibitem[\protect\citename{Cheng \bgroup et al.\egroup }2014]{cheng2014}
Cheng, J., Zhao-D'Antilio, Y., Chen, X., and Bernstein, J.
\newblock (2014).
\newblock Automatic spoken assessment of young english language learners.
\newblock In {\em Proc. of the Ninth Workshop on Innovative Use of NLP for
  Building Educational Applications}.

\bibitem[\protect\citename{Cooke}1994]{cooke1994}
Cooke, N.~J.
\newblock (1994).
\newblock Varieties of knowledge elicitation techniques.
\newblock {\em International Journal on Human-Computer Studies}, (41):801--849.

\bibitem[\protect\citename{Das and Hasegawa-Johnson}2015]{das2015}
Das, A. and Hasegawa-Johnson, M.
\newblock (2015).
\newblock Cross-lingual transfer learning during supervised training in low
  resource scenarios.
\newblock {\em Proc. of Interspeech}, pages 3531--3535.

\bibitem[\protect\citename{Das \bgroup et al.\egroup }1998]{DasNixPic98}
Das, S., Nix, D., and Picheny, M.
\newblock (1998).
\newblock {Improvements in Children's Speech Recognition Performance}.
\newblock In {\em Proc. of ICASSP}, pages 433--436, Seattle,WA, May.

\bibitem[\protect\citename{Duan \bgroup et al.\egroup }2017]{duan2017}
Duan, R., Kawahara, T., Dantsuji, M., and Zhan, J.
\newblock (2017).
\newblock Articulatory modeling for pronunciation error detection without
  non-native training data based on dnn transfer learning.
\newblock {\em IEICE Transactions on Information and Systems},
  E100.D(9):2174--2182.

\bibitem[\protect\citename{Eskenazi}2009]{eskenazi2009}
Eskenazi, M.
\newblock (2009).
\newblock An overview of spoken language technology for education. speech
  communication.
\newblock {\em Speech Communication}, 51(10):2862--2873.

\bibitem[\protect\citename{Gerosa \bgroup et al.\egroup }2007]{GerGiuBru07}
Gerosa, M., Giuliani, D., and Brugnara, F.
\newblock (2007).
\newblock Acoustic variability and automatic recognition of children's speech.
\newblock {\em Speech Communication}, 49(10-11):847 -- 860.

\bibitem[\protect\citename{Gerosa \bgroup et al.\egroup }2009]{GerGiuBru09}
Gerosa, M., Giuliani, D., and Brugnara, F.
\newblock (2009).
\newblock Towards age-independent acoustic modeling.
\newblock {\em Speech Communication}, 51(6):499 -- 509.

\bibitem[\protect\citename{Giuliani and Gerosa}2003]{GiuGer03}
Giuliani, D. and Gerosa, M.
\newblock (2003).
\newblock {Investigating Recognition of Children Speech}.
\newblock In {\em {Proc. of ICASSP}}, volume~2, pages 137--40, Hong Kong, Apr.

\bibitem[\protect\citename{Gretter \bgroup et al.\egroup }2019]{icassp2019}
Gretter, R., Matassoni, M., Allgaier, K., Tchistiakova, S., and Falavigna, D.
\newblock (2019).
\newblock Automatic assessment of spoken language proficiency of non-native
  children.
\newblock In {\em Proc. of ICASSP}.

\bibitem[\protect\citename{Lee and Glass}2015]{lee2015}
Lee, A. and Glass, J.
\newblock (2015).
\newblock Mispronunciation detection without nonnative training data.
\newblock In {\em Proc. of Interspeech}, pages 643--647.

\bibitem[\protect\citename{Li and Russell}2001]{LiRus01}
Li, Q. and Russell, M.
\newblock (2001).
\newblock {Why is Automatic Recognition of Children's Speech Difficult?"}.
\newblock In {\em Proc. of Eurospeech}, Aalborg, Denmark, Sept.

\bibitem[\protect\citename{Li \bgroup et al.\egroup }2016]{li2016}
Li, W., Siniscalchi, M., Chen, N.~F., and Lee, C.~H.
\newblock (2016).
\newblock Improving non-native mispronunciation detection and enriching
  diagnostic feedback with {DNN-based} speech attribute modeling.
\newblock In {\em Proc. of ICASSP}, pages 6135--6139.

\bibitem[\protect\citename{Liao \bgroup et al.\egroup }2015]{liao2015}
Liao, H., Pundak, G., Siohan, O., Carroll, M., Coccaro, N., Jiang, Q., Sainath,
  T.~N., Senior, A., Beaufays, F., and Bacchiani, M.
\newblock (2015).
\newblock Large vocabulary automatic speech recognition for children.
\newblock In {\em Proc. of Interspeech}.

\bibitem[\protect\citename{Matassoni \bgroup et al.\egroup }2018]{icassp2018}
Matassoni, Gretter, R., Falavigna, D., and Giuliani, D.
\newblock (2018).
\newblock Non-native children speech recognition through transfer learning.
\newblock In {\em Proc. of ICASSP}.

\bibitem[\protect\citename{Menzel \bgroup et al.\egroup }2000]{menzel2000isle}
Menzel, W., Atwell, E., Bonaventura, P., Herron, D., Howarth, P., Morton, R.,
  and Souter, C.
\newblock (2000).
\newblock The {ISLE} corpus of non-native spoken {English}.
\newblock In {\em Proc. of LREC}, pages 957--964.

\bibitem[\protect\citename{Oh \bgroup et al.\egroup }2006]{Oh2006}
Oh, Y.~R., Yoon, J.~S., and Kim, H.~K.
\newblock (2006).
\newblock Adaptation based on pronunciation variability analysis for non native
  speech recognition.
\newblock In {\em Proc. of ICASSP}, pages 137--140.

\bibitem[\protect\citename{Oh \bgroup et al.\egroup }2017]{oh2017}
Oh, Y.~R., Jeon, H.-B., Song, H.~J., Kang, B.~O., Lee, Y.-K., Park, J., and
  Lee, Y.-K.
\newblock (2017).
\newblock Deep-learning based {Automatic} {Spontaneous} {Speech} {Assessment}
  in a {Data-Driven} {Approach} for the 2017 {SLaTE} {CALL} {Shared}
  {Challenge}.
\newblock In {\em {Proc. of SlaTe}}, pages 103--108, Stockholm, Sweden.

\bibitem[\protect\citename{Potamianos and Narayanan}2003]{PotNar03}
Potamianos, A. and Narayanan, S.
\newblock (2003).
\newblock {Robust Recognition of Children's Speech}.
\newblock {\em IEEE Transactions on SAP}, 11(6):603--615, Nov.

\bibitem[\protect\citename{Qian \bgroup et al.\egroup }2019]{russell2019}
Qian, M., Jancovic, P., and Russel, M.
\newblock (2019).
\newblock The university of birmingham 2019 spoken call shared task systems:
  Exploring the importance of word order in text processing.
\newblock In {\em {Proc. of SlaTe}}, pages 11--15, Graz, Austria.

\bibitem[\protect\citename{Raab \bgroup et al.\egroup }2007]{noeth2007}
Raab, M., Gruhn, R., and Noeth, E.
\newblock (2007).
\newblock Non-native speech databases.
\newblock In {\em Proc. of ASRU}, pages 413--418, Kyoto, Japan.

\bibitem[\protect\citename{Russell}2007]{russell2007}
Russell, M.
\newblock (2007).
\newblock {Analysis of {Italian} children's {English} pronunciation}.
\newblock {\em http://archive.is/http://www.eee.bham.ac.uk/russellm}.

\bibitem[\protect\citename{Serizel and Giuliani}2016]{serizel2016}
Serizel, R. and Giuliani, D.
\newblock (2016).
\newblock Deep-neural network approaches for speech recognition with
  heterogeneous groups of speakers including children.
\newblock {\em Natural Language Engineering}, FirstView:1--26, 7.

\bibitem[\protect\citename{Steidl \bgroup et al.\egroup }2004]{Steidl2004}
Steidl, S., Stemmer, G., Hacker, C., and N\"oth, E.
\newblock (2004).
\newblock Adaptation in the pronunciation space for non-native speech
  recognition.
\newblock In {\em Proc. of ICSLP}, pages 2901--2904.

\bibitem[\protect\citename{Strik \bgroup et al.\egroup }2009]{strik2009}
Strik, H., Truong, K., {de Wet}, F., and Cucchiarini, C.
\newblock (2009).
\newblock Comparing different approaches for automatic pronunciation error
  detection.
\newblock {\em Speech Communication}, 51(10):845--852.

\bibitem[\protect\citename{Wang and Schultz}2003]{Wang2003a}
Wang, Z. and Schultz, T.
\newblock (2003).
\newblock Non-native spontaneous speech recognition through polyphone decision
  tree specialization.
\newblock In {\em Proc. of Eurospeech}, pages 1449--1452.

\bibitem[\protect\citename{Wang \bgroup et al.\egroup }2003]{Wang2003b}
Wang, Z., Schultz, T., and Waibel, A.
\newblock (2003).
\newblock Comparison of acoustic model adaptation techniques on non-native
  speech.
\newblock In {\em Proc. of ICASSP}, pages 540--543.

\bibitem[\protect\citename{Wilpon and Jacobsen}1996]{WilJac96}
Wilpon, J. and Jacobsen, C.
\newblock (1996).
\newblock {A Study of Speech Recognition for Children and Elderly}.
\newblock In {\em Proc. of ICASSP}, pages I--349--352, Atlanta, GA, May.

\bibitem[\protect\citename{Zechner and Evanini}2019]{zechner2019}
Zechner, K. and Evanini, K.
\newblock (2019).
\newblock {\em Automated Speaking Assessment: Using Language Technologies to
  Score Spontaneous Speech}.
\newblock Educational Testing Service, Princeton (NJ).

\bibitem[\protect\citename{Zechner \bgroup et al.\egroup }2009]{zechner2009}
Zechner, K., Higgins, D., Xi, X., and Williamson, D.
\newblock (2009).
\newblock Automatic scoring of non-native spontaneous speech in tests of spoken
  {English}.
\newblock {\em Speech Communication}, 51(10):883--895.

\end{thebibliography}


\end{document}